\documentclass[10pt,twocolumn,letterpaper]{article}

\usepackage{iccv}
\usepackage{blindtext}% for dummy text only
\usepackage{times}
\usepackage{epsfig}
\usepackage{graphicx}
\usepackage{amsmath}
\usepackage{amssymb}
\usepackage{caption}
\usepackage{subcaption}
\usepackage{multirow}
\usepackage{wrapfig}
\usepackage{overpic}
\usepackage{balance}
\usepackage{sidecap, caption}
\usepackage{soul}
\usepackage{enumitem}
\usepackage[ruled,linesnumbered]{algorithm2e}

\usepackage[dvipsnames]{xcolor}
\definecolor{green}{RGB}{0, 128, 0}

\definecolor{cc1}{RGB}{224, 62, 30}
\definecolor{cc2}{RGB}{254, 209, 48}
\definecolor{cc3}{RGB}{7, 124, 219}
\definecolor{cc4}{RGB}{102, 204, 0}

\usepackage{booktabs}
\newcommand{\ra}[1]{\renewcommand{\arraystretch}{#1}}

\usepackage[breaklinks=true,bookmarks=false,colorlinks=true]{hyperref}

\iccvfinalcopy % *** Uncomment this line for the final submission

\def\iccvPaperID{ciao} % *** Enter the ICCV Paper ID here

\ificcvfinal\fi

\begin{document}

%%%%%%%%% TITLE
%\title{Visual and Semantic Classifier Crafting: \\ Turn Your ConvNet into an (Interpretable) Zero-Shot Learner!}
\title{Classifier Crafting: Turn Your ConvNet into a Zero-Shot Learner!}

\author{Jacopo Cavazza\\
Istituto Italiano di Tecnologia \\
Pattern Analysis and Computer Vision (PAVIS) Department\\
Visual Geometry and Modelling (VGM) \\
GREAT Campus: Science and Technology Park in Genoa\\Via Enrico Melen, 83, 16152 Genoa GE, Italy\\
{\tt\small jacopo.cavazza@iit.it}
% For a paper whose authors are all at the same institution,
% omit the following lines up until the closing ``}''.
% Additional authors and addresses can be added with ``\and'',
% just like the second author.
% To save space, use either the email address or home page, not both
}

\twocolumn[{%
\maketitle
\renewcommand\twocolumn[1][]{#1}%
\begin{center}
    \centering
    \vspace{-15pt}
    \begin{overpic}[width=0.8\textwidth]{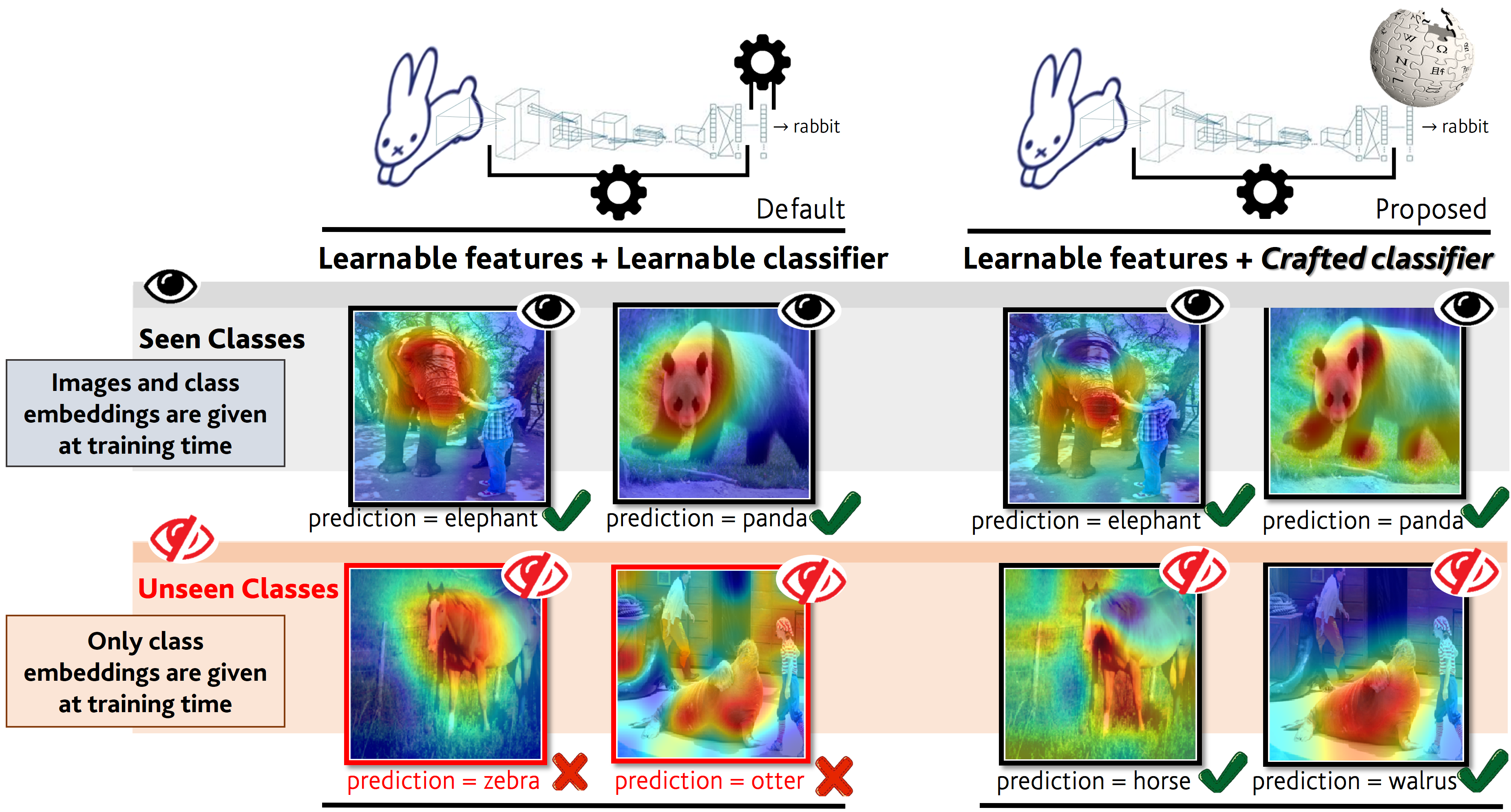}
    \end{overpic}
    \captionof{figure}{{\bf The idea.} We propose a learnable representation, tailored for ZSL, obtained from a convolutional network in which we \emph{crafted} the softmax classifier's weights. As opposed to optimize these weights (\protect\includegraphics[width=0.35cm]{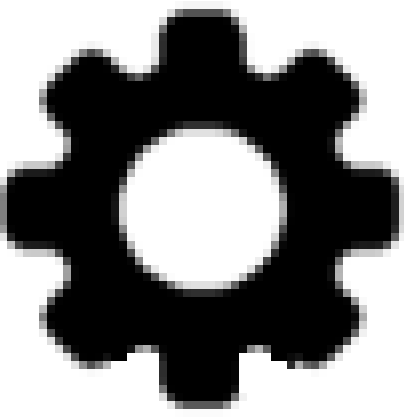}), %In a data-driven softmax operator, its weights are optimized with training data: hence, they are arguably not able to cope with unseen classes at inference time. Differently, we \textit{craft} the weights, 
    we use \textit{fixed classification rules}: \eg, we replace the softmax weights' with class embeddings, by either leveraging Osherson's default probability scores \cite{lampert2009learning} or LSTM-CNN text embeddings trained on Wikipedia \cite{xian2018zero}. Then, we learn a feature representation to match these fixed classification rules: we posit this to be the proxy to turn a ``vanilla'' convnet, trained on seen images only, into a zero-shot learner. Further, we achieve interpretability at no additional cost, using neural attention (\eg, grad-CAM \cite{selvaraju2017grad}): our ZSL-crafted nets seem to focus on finer semantic details (\eg, legs or proboscis when classifying animals) and to better deal with confounding factors (\eg, in this case, a person).}%%by crafting the weights of the softmax classifier of a standard convnet through a pool of fixed classification rules, that generalize by design towards unobserved classes. Here, we crafted the softmax of a ResNet-101 \cite{he2016deep} model (although in this study we also consider AlexNet \cite{krizhevsky2012imagenet}, DenseNet-201 \cite{huang2017densely} and DarkNet-53 \cite{redomon2020darknet}) using class embeddings of \emph{seen} classes (\eg, ``elephant'') from Animals with Attributes 2 \cite{lampert2009learning}. The net was then fine-tuned on \emph{seen} training data only. Since class embeddings are given also for  \emph{unseen} classes (\eg, ``horse''), we can recognize them injecting their respective classification rules into the classifier, performing inference over unseen (and seen) classes using expanded set of logits (Algorithm \ref{alg:crafting}). Re-training on unseen data is totally avoided. This turns a vanilla convnet into a ZSL predictor with a feature representation that is ZSL-specific and learnt from (seen) images directly (as opposed to be pre-computed for ImageNet classification \cite{xian2019f-VAEGAN-D2,schonfeld2019generalized,Huang_2019_CVPR,li2019leveraging,xu2020attribute,verma2020meta,keshari2020generalized,chou2021adaptive}). Ancillary, we gain interpretability at no additional cost: using grad-CAM \cite{selvaraju2017grad}, we discover that our crafted ResNet-101 shows a more focalized attention on specific body parts, like eyes and legs (\emph{bottom-right pane}) or ears and proboscis (\emph{top-right pane}). This is opposed to a generically spread attention (of a standard, vanilla ResNet-101) which keep focusing on the whole animal's head (\emph{bottom-left pane}) or on task-irrelevant elements (\eg, here, a person, \emph{top-left pane}). %We posit that better grounding semantic cues into the visual representation is a desirable feature of our proposed crafting, in which, fine-tuning using transferable classification rules results in visual representations which are tailored for zero-shot learning, hence better (Table \ref{tab:soa}) as opposed to use pre-trained ones as in prior art     \cite{xian2019f-VAEGAN-D2,schonfeld2019generalized,Huang_2019_CVPR,li2019leveraging,xu2020attribute,verma2020meta,keshari2020generalized,chou2021adaptive}. 
    \label{fig:sell}
\end{center}%
}]%

%\maketitle

% Remove page # from the first page of camera-ready.
\ificcvfinal\thispagestyle{empty}\fi

%%%%%%%%% ABSTRACT
\begin{abstract}
In Zero-shot learning (ZSL), we classify unseen categories using textual descriptions about their expected appearance when observed (\textit{class embeddings}) and a disjoint pool of seen classes, for which annotated visual data are accessible. We tackle ZSL by casting a ``vanilla'' convolutional neural network (\eg AlexNet \cite{krizhevsky2012imagenet}, ResNet-101 \cite{he2016deep}, DenseNet-201 \cite{huang2017densely} or DarkNet-53 \cite{redomon2020darknet}) into a zero-shot learner. We do so by crafting the softmax classifier: we freeze its weights using fixed seen classification rules, either semantic (seen class embeddings) or visual (seen class prototypes). Then, we learn a data-driven and ZSL-tailored feature representation on seen classes only to match these fixed classification rules. Given that the latter seamlessly generalize towards unseen classes, while requiring not actual unseen data to be computed, we can perform ZSL inference by augmenting the pool of classification rules at test time while keeping the very same representation we learnt: nowhere re-training or fine-tuning on unseen data is performed. The combination of semantic and visual crafting (by simply averaging softmax scores) improves prior state-of-the-art methods in benchmark datasets for standard, inductive ZSL. After rebalancing predictions to better handle the joint inference over seen and unseen classes, we outperform prior generalized, inductive ZSL methods as well. Also, we gain interpretability at no additional cost, by using neural attention methods (\eg, grad-CAM \cite{selvaraju2017grad}) as they are. Code will be made publicly available.
\end{abstract}
%We turn these convnets into zero-shot learners by \emph{crafting} the classifiers' weight which are frozen to a fixed configuration (using either class embeddings or learnt visual prototypes for that), while other layers are fine-tuned to accommodate for the change. At the inference stage, we augment the classifier's weights %Often, doing so on the penultimate layer is sufficient and, to the extreme, the fine-tuning can be totally avoided since even just remapping pre-computed features onto attributes is effective. When combined with an ensemble and a predictions' rebalancing mechanisms, our proposed crafting improves upon state-of-the-art methods on standard and generalized inductive ZSL benchmark datasets. Furthermore, our idea allows for using popular attentional maps for convnets (\eg, \cite{selvaraju2017grad}) to gain interpretability for free.

%%%%%%%%% BODY TEXT

\section{Introduction}

\begin{figure}[t!]
    \centering
    \includegraphics[width = \columnwidth]{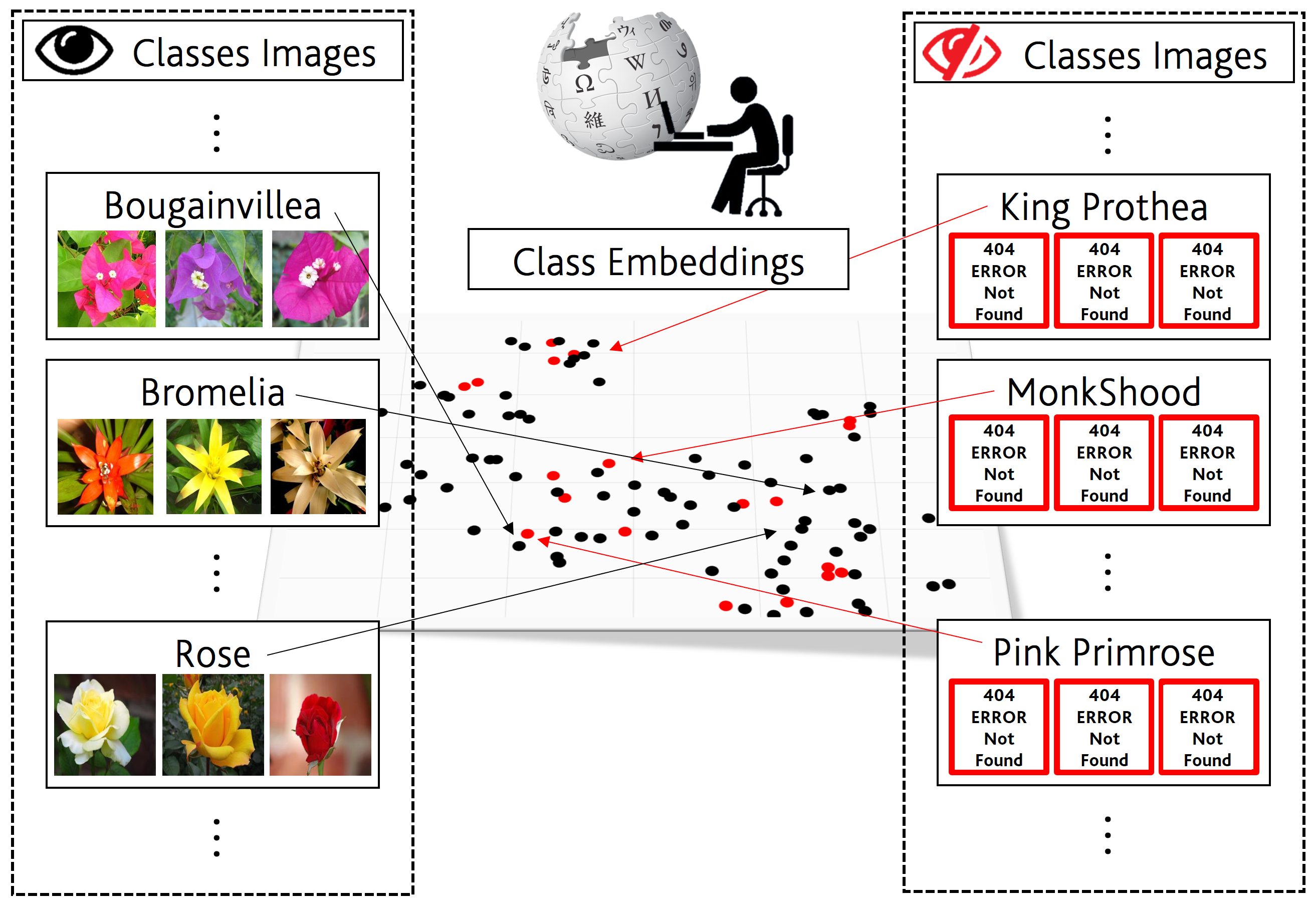}
    \caption{In this paper, we consider the problem of {\bf inductive zero-shot learning}, considering both standard and generalized evaluation paradigms, recognizing unseen classes (no visual data at training time) using class embeddings (\eg, a list of attributes) and seen categories (provided of annotated visual data).}%{\bf Standard and generalized, inductive ZSL}. Seen classes are provided of annotated visual data. Unseen classes are disjoint from the previous ones and totally deprived of visual data (we \textit{do not} consider here the easier \textit{transductive} variant \cite{xian2018zero}). Both seen and unseen classes are described with auxiliary information termed \emph{class embedding}: these are vectorial numerical representations (one per class) are obtained by either manually defined-attributes (tabulated into float numbers through Osherson's scores \cite{lampert2009learning}) or distributed word embeddings (\eg, computed from a LSTM model parsing Wikipedia articles \cite{xian2019f-VAEGAN-D2}). In \textit{standard, inductive ZSL}, inference is done on unseen clases only, while \textit{generalized, inductive ZSL (GZSL)} jointly considers seen and unseen classes at test time.}
    \label{fig:ZSL_problem}
\end{figure}

Due to the intrinsic difficulty of gathering a sufficient number of annotations for visual categories worth to be recognized, inductive zero-shot learning (ZSL) is advantageous since tackling the ``extreme'' scenario where some of the classes are never observed by the learner at training time (\textit{unseen classes}), but still predicted at inference time (see Fig. \ref{fig:ZSL_problem}). In fact, unseen classes can be recognized by exploiting textual descriptions (dubbed \textit{class embeddings}) that inform the learner about how unseen classes are supposed to appear when observed \cite{lampert2009learning,chao2016empirical}. In conjunction to this, a disjoint pool of categories (\textit{seen classes}) is accessible in terms of both annotated visual data (here, images) and, again, class embeddings. In abstract terms, ZSL it's about relating seen class embeddings (and their semantic content) with visual seen data by means of a mapping which, despite being optimized on seen data only, has to \textit{transfer} over unseen classes (standard ZSL) and over both unseen classes and unseen instances from seen classes (generalized ZSL, GZSL).

Previously, the way of making the aforementioned ``mapping'' transferrable, has been mainly tackled from the algorithmic perspective (either by finding a mathematical alignment between visual and semantic domains - \cite{xian2018zero} for a review - or training a feature generator to synthesize visual features from class embeddings \cite{xian2019f-VAEGAN-D2,schonfeld2019generalized,Huang_2019_CVPR,li2019leveraging,xu2020attribute,verma2020meta,keshari2020generalized,chou2021adaptive}) but almost always relying on pre-computed visual features that are extracted from ImageNet pre-trained neural networks \cite{xian2018zero,xian2019f-VAEGAN-D2,schonfeld2019generalized,Huang_2019_CVPR,li2019leveraging,xu2020attribute,verma2020meta,keshari2020generalized,chou2021adaptive}.

\paragraph{Contributions.} In this paper, we propose to entangle the problem of making the aforementioned ``seen-to-unseen mapping'' generalizable with the problem of learning ZSL-specific visual descriptors.  We \emph{craft} the weights $w_i$ of the softmax classifier's of a standard convnet: that is, we keep $w_i$ freezed and we optimize by gradient descent the rest of the convnet accordingly. We propose two strategies for that. 1) \textit{Semantic crafting}: we replace $w_i$ with the class embeddings which, thanks to Osherson's default probabilities\footnote{\textit{E.g.}, we will give probability ``1'' for a zebra to be four legged and probability ``0'' for a flying otter. Note that these probabilities are not visually grounded \cite{bustreo2019enhancing} and only convey information on the expected appearance: not all images of a zebra will actually show four legs!} \cite{lampert2009learning} or CNN-LSTM text embeddings \cite{xian2019f-VAEGAN-D2} can be casted into numerical vectors (that's what $w_i$ are). 2) \textit{Visual crafting}: we replace $w_i$ with visual prototypes. While seen prototypes are simply obtained by averaging visual features, unseen prototypes are inferred through a linear projection (optimized in closed-form) from class embeddings, in order not to violate the ZSL assumption. In both cases, we fine-tune the hierarchical feature representation of the convnet \textit{using seen images only}, once the classifier is frozen after it has been visually or semantically crafted. Since our classification rules naturally extend towards unseen classes (albeit not requiring unseen data to be computed), we claim that the combination of them with a learnable and ZSL-tailored representation is a valuable proxy to generalize over unseen classes while being able to fine-tune a learnable representation only on the seen classes.

%Since softmax predictions mathematically write as the inner product between the (highest-level) feature representation $\mathbf{f}$ and the softmax's weights $w_i$, at the inference stage, we augment $w_i$ with a new pool of weights $\widetilde{w}_k$, where $\widetilde{w}_k$ is known one-to-one correspondence to an unseen class. We can do so since class embeddings are disclosed even for unseen classes at training time (sematic crafting) and visual prototypes for unseen classes are inferred using the projection learnt using seen data only (visual crafting): in both cases, we are consistent with the inductive ZSL assumption. As the result, we can perform inference over unobserved categories despite fine-tuning is done on seen data only. The reason is that our fixed classification rules intrinsically generalize towards unseen classes and, arguably, the same property is inherited from the visual representation we optimize to fit these rules.

%unseen classes and obtained by either unseen class embeddings or estimated unseen prototypes (using the mapping we learnt on the seen data only). The softmax score is the computed using both seen $w_i$ and unseen $\widetile{w}_j$ weights, but no re-training on the unseen classes is performed.

%simply re-training a newly added layer to match the dimensions of the class embeddings and the feature representation $\mathbf{f}$ fed to the softmax classifier. Second, we start from a pre-trained network, we compute \emph{seen} visual prototypes (by averaging the high-level feature representation $\mathbf{f}$

Since visual and semantic data are complementary in nature, we combine these two crafted convnets in an ensemble by simply averaging their predictions obtained from visually- and semantically-crafted softmax operators: in this manner, we improve prior art in standard, inductive ZSL \cite{xian2019f-VAEGAN-D2,schonfeld2019generalized,Huang_2019_CVPR,li2019leveraging,xu2020attribute,verma2020meta,keshari2020generalized,chou2021adaptive} -- see Table \ref{tab:soa}. 

To better deal with the GZSL setup, we propose a novel confidence rebalancing mechanism to fill the gap between seen predictions and (arguably) under-confident unseen predictions, given that unseen images are totally unobserved at training time. We do so by learning a discriminator to predict whether a vector of logits is seen (or not) using auxiliary task-irrelevant data to optimize it. Afterwards, we can use it to re-modulate, on the fly, the seen and unseen softmax probabilities at test time before taking a decision (again, neither training nor fine-tuning on unseen data is performed). By increasing the confidence of unseen classes and decreasing the ones of seen classes, (see Fig. \ref{fig:pipeline_inference}), our rebalanced ensemble outperforms prior art in generalized, inductive ZSL \cite{xian2019f-VAEGAN-D2,schonfeld2019generalized,Huang_2019_CVPR,li2019leveraging,xu2020attribute,verma2020meta,keshari2020generalized,chou2021adaptive} -- see Table \ref{tab:soa}. 

We dissect the impact in performance of the visual and semantic crafting strategies, the rebalancing mechanism and the ensemble in ablation studies. We also provide grad-CAM \cite{selvaraju2017grad} visualizations to gain in interpretability at no additional cost (see Fig. \ref{fig:sell}).

%{\bf Contributions}

%\begin{itemize}
%\item We extend fixed Classifiers for ZSL: take a net, replace the classifier, fine-tune and ciao! 
%\item We propose two novel zero-shot learners based on this idea, whose prediction are compatible through an ensemble mechanism.
%\item We propose a novel recalibration technique to jointly resolve over- and under-confidence of the ZSL learner over seen and unseen classes, respectively, ultimately easing the generalized zero-shot learning task. This techniques, based on auxiliary unseen data and mixup \cite{zhang2018mixup}, outperforms existing analogous solutions \cite{chao2016empirical,atzmon2019COSMO}.
%\item Several ablation studies, while also providing a study on interpretation of the scored results, borrowing the existing weakly supervised saliency map frameworks which can be applied to a zero-shot learner thanks to our idea.
%\item We provide an experimental result over several benchmark dataset, scoring a favorable performance in terms in improvements over sota
%\end{itemize}

To summarize, these are the main contribution of our work:
\begin{itemize}[leftmargin=*]
    \item We propose (visual and semantic) fixed classification rules to craft (\ie, replace and keep freezed) the softmax classifier's weights. This turns a convnet into a zero-shot learner, Consequently, we learn \textit{from seen visual data only} a feature representation which is tailored for ZSL and, thus, effective to recognize unseen classes as well.
    \item With a simple ensemble method (average of predictions) our visual and semantic crafted networks improve prior art in standard, inductive ZSL. Once endowed of a novel predictions' rebalancing mechanism, which we propose, our crafted networks improve improve prior art in inductive GZSL as well.
    \item We provide ablation studies to quantify the role of each computational module in performance, comparing with prior rebalancing mechanisms \cite{chao2016empirical} and \cite{atzmon2019COSMO}. We also gain interpretability simply using grad-CAM \cite{selvaraju2017grad} as it is.
\end{itemize}

\paragraph{Outline of the paper.~~} We first explain the proposed approach (Section \ref{sez:method}) and present the supporting experimental evidences (Section \ref{sez:exp}). The discussion of related work is deferred to Section \ref{sez:relwork}, in order to, hopefully, gain in readability. Conclusions will be drawn in Section \ref{sez:conclusions}.

\section{Method}\label{sez:method}

%We operate in the inductive ZSL scenario (in both its standard and generalized variants) depicted in %in which two set of classes are available: \emph{seen} and \emph{unseen}. Seen classes are paired of visual data (in this paper, images) which are, instead, totally missing for unseen ones (see ...) Fig. \ref{fig:ZSL_problem}. 
For seen classes, input data are triplets of image, class label and a class embedding (the latter being a numerical vector in one-to-one known correspondence to a class). Unseen classes are totally deprived of images and are known only in terms of their class ebmeddings: unseen visual data are disclosed only at test time and no re-training on them is allowed.

\begin{figure}[h!]
    \centering
    \includegraphics[width =0.5\columnwidth]{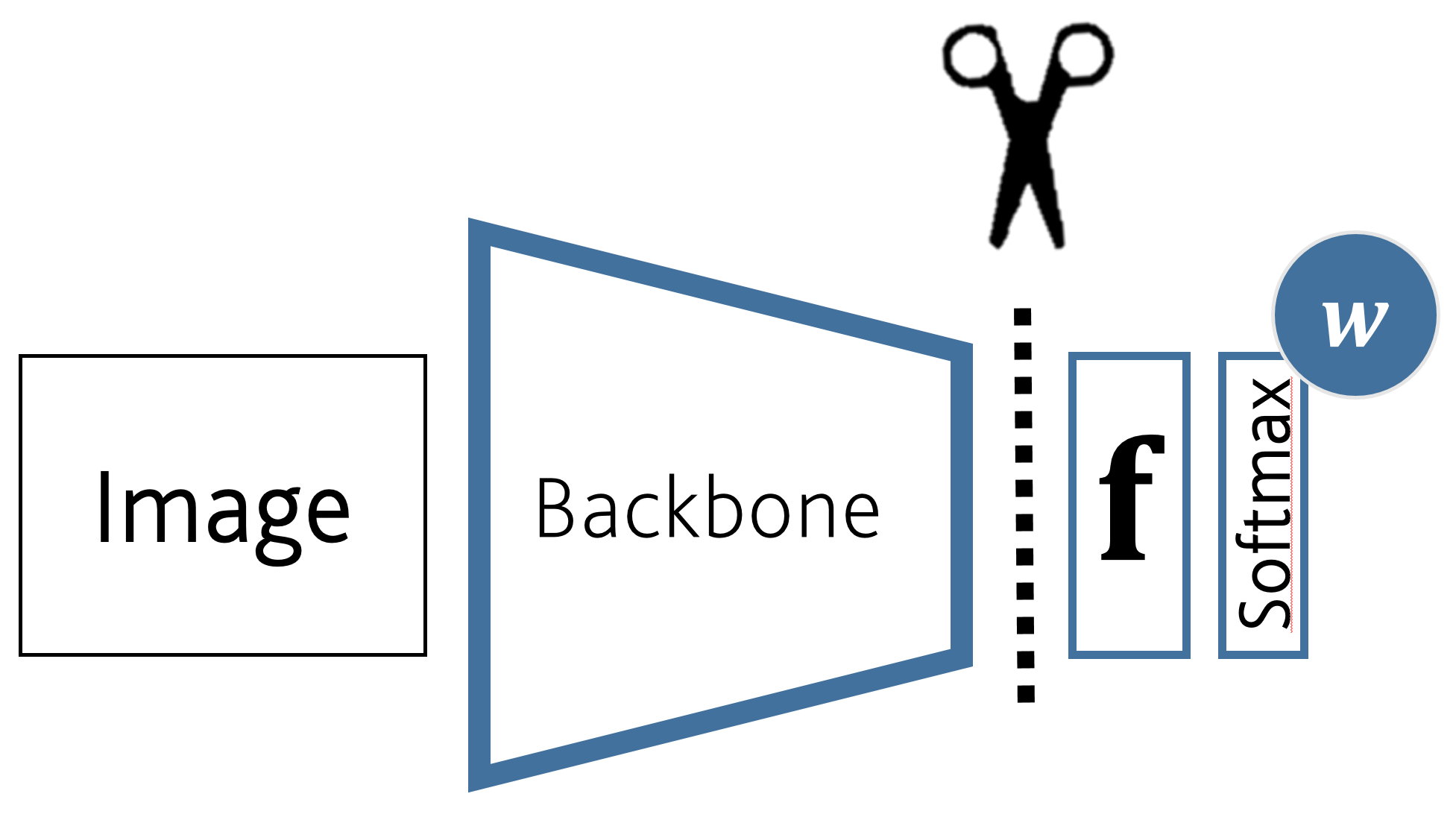}
    \caption{Crafting the softmax's weights $w_j$ of a convnet: we fine-tune the representation $\mathbf{f}$ while freezing $w_j$.}
   \label{fig:CNN_pre_trained}
\end{figure}

%Let us here clarify the difference between standard and generalized evaluation protocols in inductive ZSL: in the former, which we tag as ZSL if confusion does not arise, performance is evaluated over unseen classes only using top-1 classification accuracy $T1$. In the generalized ZSL protocol, usually tagged as GZSL, inference is done on either seen and unseen classes and the recognition capability is evaluated on the two groups separately. To this aim, we take advantage of $S$ and $U$, defined as mean per-class classification accuracy over seen and unseen classes, respectively \cite{xian2018zero}. To aggregate these two metrics, their harmonic mean $\mathcal{H}$ \cite{xian2018zero} of the area under the seen and unseen curve $\mathcal{A}$ \cite{atzmon2019COSMO} is used for evaluation. \textcolor{red}{See the Supplementary Material for a rigorous mathematical definition of $S,U,\mathcal{H}$ and $\mathcal{A}$.}
%Class embeddings are instead, given for both seen and unseen categories: denoted by $\mathbf{a}$, class embeddings are usually annotated by humans in the form of Osherson's scores, measuring the abstract compatibility of an attribute with a category (like ``is quadrupedal'' for a ``horse'') \cite{xian2018zero}. Alternatively (\eg, see \cite{xie2021cross}), distributed word embeddings or end-to-end NLP architectures  can be used to learn how to embed (the name of) a category into a vector such that semantical similarity translates into small (Euclidean) distance.

\begin{figure*}
    \centering
         \begin{minipage}[b]{0.49\textwidth}
         \centering \includegraphics[width=\textwidth]{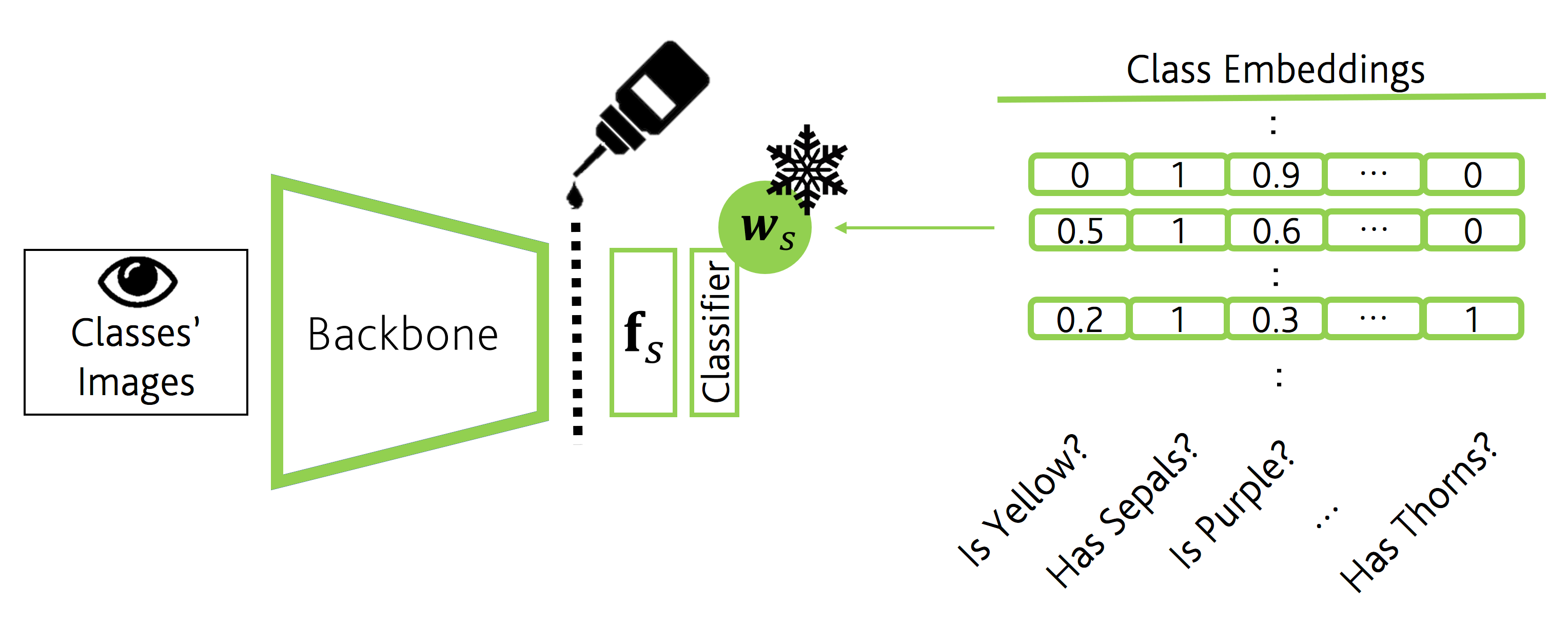}
         %\caption{S-CC: Crafting $\mathbf{w}_s$ in the semantic space}
         %\label{fig:FW_semantic}
     \end{minipage}
         \begin{minipage}[b]{0.49\textwidth}
         \centering
         \includegraphics[width=\textwidth]{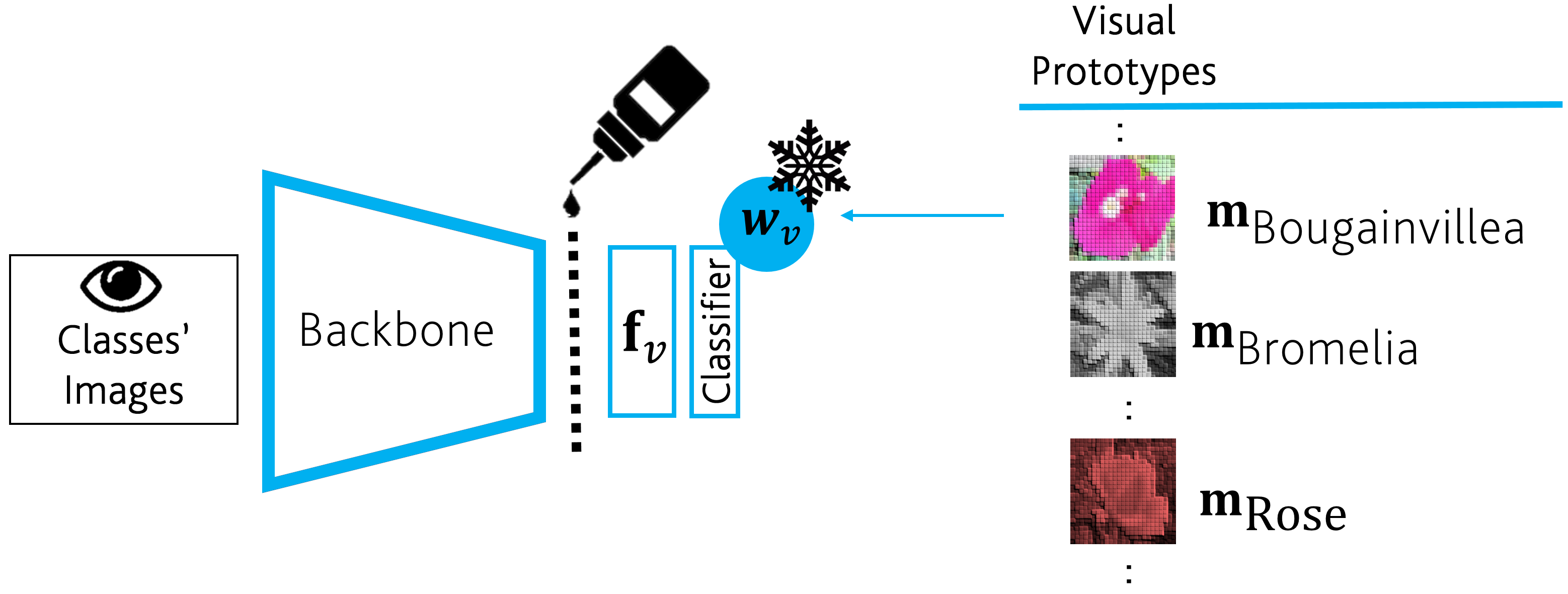}
         %\caption{V-CC: Crafting Weights $\mathbf{w}_v$ in the visual space}
         %\label{fig:FW_visual}
     \end{minipage}
     \caption{{\bf S-CC} (\emph{left}) and {\bf V-CC} (\emph{right}): Semantic and Visual Classifier Crafting using fixed, transferrable classification rules.}
     \label{fig:pipeline_training}
\end{figure*}

%While doing inference, in standard ZSL (referred as ZSL if confusion does not arise), performance is evaluated in terms of top-1 accuracy $T1$ over the unseen classes only. In generalized ZSL, instead, inference has to be jointly done on test examples from both seen and unseen classes: the metrics $S$ and $U$ compute the mean average per class accuracy over seen and unseen classes, respectively (\textcolor{red}{formulas in the Supplementary Material}). The final generalized ZSL performance is evaluated either in terms of $H$ (harmonic mean between $S$ and $U$ \cite{xian2018zero}) in terms of $\mathcal{A}$, the area under the seen-and-unseen curve \cite{atzmon2019COSMO}. \textcolor{red}{da capire bene come calcolarla, dovrebbe essere cv di calibrate stacking e poi trapezi per integrare.}

%\begin{wrapfigure}{r}{0.45\columnwidth}
%\end{wrapfigure} 

\paragraph{Backbone Network.} We consider a convolutional feed-forward neural network (convnet) with a softmax classifier, trained in and end-to-end on raw images. Using the softmax, inference over a fixed pool of classes (indexed over $j$), can done through $\arg \max_j \mathbf{p}_j$ where
\begin{equation}\label{eq:softmax}
    \mathbf{p}_j = \frac{\exp( \mathbf{f} \cdot w_j) }{\sum_\ell \exp( \mathbf{f} \cdot w_\ell)}
\end{equation}
is the probability of predicting the $j$-th class\footnote{For efficiency issues, $\arg \max_j \mathbf{p}_j$ is usually replaced by $\arg \max_j \mathbf{f} \cdot w_j$ in standard deep learning libraries (\eg, PyTorch, MATLAB, Tensorflow, \dots), since both operations give the same result.}. In Eq. \eqref{eq:softmax}, the classifier's weights are denoted by $w_j,\dots,w_\ell,\dots$  (one weight $w_j$ per class) and $\mathbf{f}$ refers to the higher lever feature representation that is learnt (and fed to the classifier to compute the logits $\mathbf{f} \cdot w_j$). 

%A vanilla convnet, pre-trained on ImageNet, is usually adopted in prior ZSL art \cite{li2019leveraging,xian2019f-VAEGAN-D2,verma2020meta,xu2020attribute,keshari2020generalized,shen2020invertible,narayan2020latent,liu2020zero,sko2021norm,chou2021adaptive} to extract visual features that are then, fed to an independent ZSL module. 
%which builds on the design of ZSL seen/unseen class splits in benchmark datasets which are compatible to ImageNet pre-training:%\begin{figure}[h!] the proposed splits ``PS'' by \cite{xian2018zero}.
%Differently, in our case, we propose to turn such a convnet into a zero-shot learner through the following two (semantic and visual) \textit{crafting mechanisms}. 

\subsection{Semantic and Visual Classifier Crafting}\label{sez:cc}

We \textit{craft} the softmax weights, that is, we replace their data-driven optimization by gradient descent with classification rules $r$ kept fixed during training. By ``classification rule'' $r$, we mean vectorial embeddings to be used to replace the weights $w_j$ in Eq. \eqref{eq:softmax}. Classification rules are ``fixed'': gradient descent is done to optimize $\mathbf{f}$ (and the whole backbone network of Fig. \ref{fig:CNN_pre_trained}) to match them, while we keep them freezed. The rationale of doing this is the following: first, we endow our fixed classification rules of semantic patterns that seamlessly generalize onto unseen classes. That is, we either exploit (continuous) class embeddings - which we have for both seen and unseen classes - or visual prototypes - computed by averaging seen descriptors or using a semantic-to-visual mapping to infer unseen ones. Second, we learn a feature representation to match these fixed classification rules: this is the proxy we propose to adopt to generalize onto unseen classes, while being able to train on seen only.

% %% qui
% Albeit the fixed softmax weights were already explored in either fully supervised vanilla classification \cite{hoffer2018fix,pernici2019maximally}, incremental learning \cite{pernici2020class} and few-shot learning \cite{qi2018low}, to the best of our knowledge, we are the first to apply this idea to inductive zero-shot learning. Differently to \cite{hoffer2018fix,pernici2019maximally,pernici2020class,qi2018low}, we are not allowed to use even a single instance of some of the classes to recognize (the unseen), but we're still asked to generalize on them. 

% To accommodate for the lack of unseen data, we exploit fixed classification rules that depends upon class embeddings (fixed to be, as said, either Osherson's probability scores or CNN+LSTM text embeddings). Even if \textit{we use only seen data for gradient descent at training time, we purport that our learnt feature representation $\mathbf{f}$ will seamlessly transfer onto unseen classes, since class embeddings own such a peculiar property by design and we optimize $\mathbf{f}$ to match class embeddings}.   

As shown in Algorithm \ref{alg:crafting}, we perform inference by computing the inner product between $\mathbf{f}$ and the classification rules $r$. Seen classification rules $r_i$ are used to learn $\mathbf{f}$. Crucially, we can still use $\mathbf{f}$ to recognize unseen classes as well, by augmenting the pool of seen logits $\mathbf{f} \cdot r_i$ with additional ones: we compute the inner product $\mathbf{f} \cdot \widetilde{r}_k$ between the very same representation $\mathbf{f}$ we learnt from seen data (and now kept freezed at inference time) and $\widetilde{r}_k$. $\widetilde{r}_k$ is a generic unseen classification rule. Thus, upon augmenting $\mathbf{f} \cdot r_i$ with $\mathbf{f} \cdot \widetilde{r}_k$, we can perform inference over seen and unseen classes jointly, with a single $\arg\max$ pass over 
%for which we know the class embedding by simply using unseen class embeddings to build unseen fixed classification rules $\widetilde{r}_k$. Hence, we simply compute seen logits $\mathbf{f} \cdot r_i$ for every seen class $i$ and unseen logits $\mathbf{f} \cdot \widetilde{r}_k$ for every unseen class $k$, predicting the most likely class by doing $\arg \max$ over seen and unseen logits 
$\mathbf{f} \cdot r_i, \dots, \mathbf{f} \cdot \widetilde{r}_k, \dots$.

 \begin{algorithm}[t!]
    \SetAlgoLined
    \KwData{Fixed rules: $r_i$ (seen) and $\widetilde{r}_k$ (unseen).}\vspace{7pt}
    %\KwResult{how to write algorithm with \LaTeX2e }
    $\pmb{\rm Training}$ (only using seen rules $r_i$ and seen data)
    \hrule
    \vspace{2pt}
    Replace the softmax weights with $r_i$ \;
    \While{not converged}{
        Keep $r_i$ frozen and fine-tune the backbone\;
        }\vspace{7pt}
    $\pmb{\rm Testing}$ 
    \hrule
    \vspace{2pt}
    \ForEach{test image}{
    Forward the test image in the backbone $\longrightarrow \overline{\mathbf{f}}$\;
    Compute logits $\overline{\mathbf{f}} \cdot r_i,\dots,\overline{\mathbf{f}} \cdot \widetilde{r}_k,\dots$ \;
    Predict the class of maximal logit\;
    }
\caption{Classifier Crafting}
\label{alg:crafting}
\end{algorithm}

\noindent \textit{Semantic Classifier Crafting $\colon$ \textbf{S-CC}.} (Fig. \ref{fig:pipeline_training} - \emph{left pane}). We select the fixed classification rules $r_i$ and $\widetilde{r}_k$, for seen classes $i$ and unseen classes $k$, to be equal to the class embeddings.

\noindent \textit{Visual Classifier Crafting $\colon$ \textbf{V-CC}} (Fig. \ref{fig:pipeline_training} - \emph{right pane}). We select $r_i$ and $\widetilde{r}_k$ as tvisual prototypes. Seen prototypes $\mathbf{m}_i$ are computed by a plain average of seen training features. Unseen prototypes $\widetilde{\mathbf{m}}_k$ are inferred using a linear projection, the latter being trained to map seen class embeddings onto seen prototypes. Please note that not a single unseen visual features is used to compute $\widetilde{\mathbf{m}}_k$, $k$ being a generic unseen class. %Additional details in the Supplementary Material.

\noindent \textit{Ensemble mechanism $\colon$ \textbf{V\&S-CC}.} Given the arguable complementarity of semantic and visual crafting, we propose a simple ensemble strategy which is based on averaging the softmax scores of a S-CC an V-CC model. Although arguably simple, this method is capable of outperforming prior art in standard, inductive ZSL (Table \ref{tab:soa}, left columns) showing the benefit of learning a tailored representation for ZSL from raw image data, as opposed to build upon pre-computed visual embeddings. %auxiliary textual descriptions and visual data in ZSL, we conjecture the S-CC and V-CC models to be complementary given the semantically- and visually-driven crafting mechanisms that we applied. Therefore, we combine those two models with a plain ensemble mechanism, obtaining by averaging their softmax probabilities, smoothened using a temperature parameter $\tau$, before performing $\arg \max$ over classes. Note that we can optionally insert the rebalancing mechanism before averaging (depending on whether we tackle standard, inductive zero-shot learning - no rebalancing is necessary - or generalized, inductive zero-shot learning).

\sidecaptionvpos{figure*}{t}
\begin{SCfigure*}[][t!]
\includegraphics[width = 0.5\textwidth]{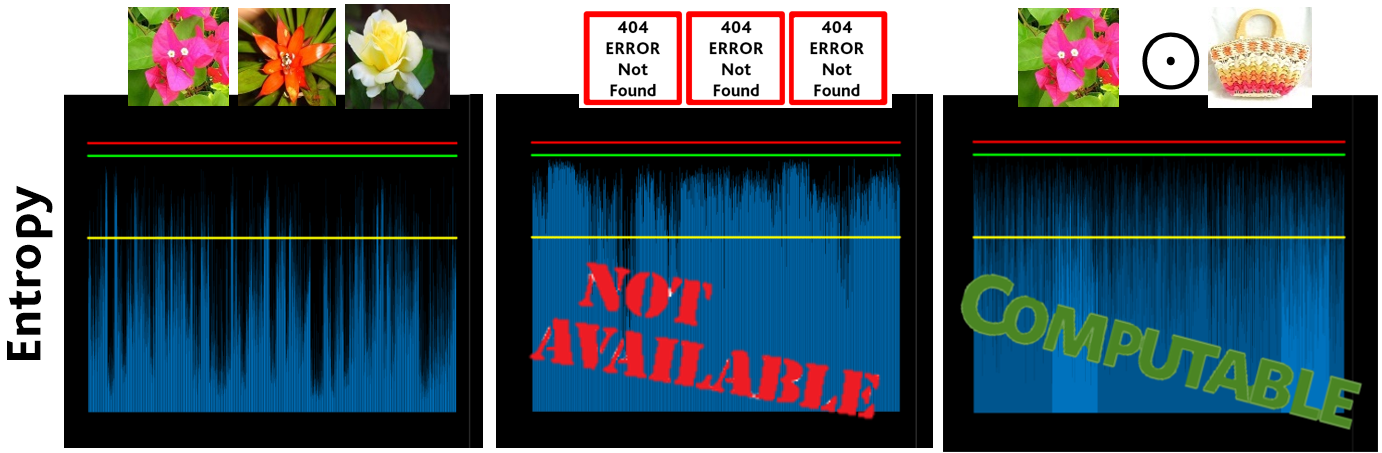}
\caption{{\bf Rebalancing Predictions}. We train a discriminator $D$ (Section \ref{ssez:rebal}) to predict if a logits is seen or not. The positive data are the seen train logits (whose entropy is visualized in the \textit{left pane}). The negative data are the mixup \cite{zhang2018mixup}, denoted by $\odot$, between seen logits and task-irrelevant data: here, ``bags'' when classifying flowers (\textit{right pane}). Thus, we generate synthetic logits that visually ``simulates'' the entropy (and thus, the uncertainty) of the unseen logits which we cannot access during training (\textit{middle pane}).}
\label{fig:rebalance}
\end{SCfigure*}

\subsection{GZSL: rebalancing seen \& unseen predictions}\label{ssez:rebal}

%\paragraph{Rebalancing predictions : S-CC-R, V-CC-R and V\&S-CC-R}. 
%Our proposed crafting-based solution for ZSL is a fully discriminative approach: since no feature generation is performed, prior art warns us about a potential limitation in  performing sufficiently well in generalized zero-shot learning \cite{xian2019f-VAEGAN-D2,schonfeld2019generalized,Huang_2019_CVPR,li2019leveraging,xu2020attribute,verma2020meta,keshari2020generalized,chou2021adaptive}. Since we empirically verify that, the performance in standard ZSL is extremely solid (better than prior art, Table \ref{tab:oracle}) we conjecture that we can opt for an alternative solution to generative approaches (which unfortunately, does not generate images for unseen classes, but only features - and we do need images to be fed into our network). 

Given the unavailability of unseen data at training time, fine-tuning our crafted network on seen data only will inherently bias the network towards them. This is expecting to result in much lower confidence for unseen classes if compared to the seen ones. This is surely a problem to tackle, since limits generalization: prior methods mainly approached this through unseen feature generation \cite{xian2019f-VAEGAN-D2,schonfeld2019generalized,Huang_2019_CVPR,li2019leveraging,xu2020attribute,verma2020meta,keshari2020generalized,chou2021adaptive}, but we propose to follow the perspective of \cite{chao2016empirical} and \cite{atzmon2019COSMO} and try to resolve the problem in the prediction space, by rebalancing the network's confidence - so that the we can increase the one for unseen classes and jointly decrease the one for seen ones.

%We build on the evidence that our ensemble of our semantically and visually crafted convnets outperforms prior art in standard, inductive ZSL (Table \ref{tab:soa}, left columns). In fact, we can always remap standard, inductive ZSL into generalize, inductive ZSL by assuming the presence of an oracle which would help us in predicting, for any test instance to be classifier, its expected membership to the pool of seen or unseen classes. If we were provided of such orcale, we can be confident in effectively tackling the generalized, inductive ZSL problem: since seen data are available, fitting them using an (over-parametrized) convent seems surely not problematic. Concurrently, however, the prediction for unseen classes (when they are exclusively considered) seems optimal, furthermore proving that our idea of crafting a classifier can really turn a convent into an effective zero-shot learner. 

In inductive GZSL, a prickly problem is mis-classifying an unseen class as if it was seen \cite{chao2016empirical}. As faced in \cite{chao2016empirical} and \cite{atzmon2019COSMO}, we can formalized it as "finding the best approximation for an oracle which, for every test instance (either seen or unseen) to be classifier, is able to tell whether the instance was sampled from the seen classes or not". 

\begin{figure*}[t!]
    \centering
    \includegraphics[width = \textwidth]{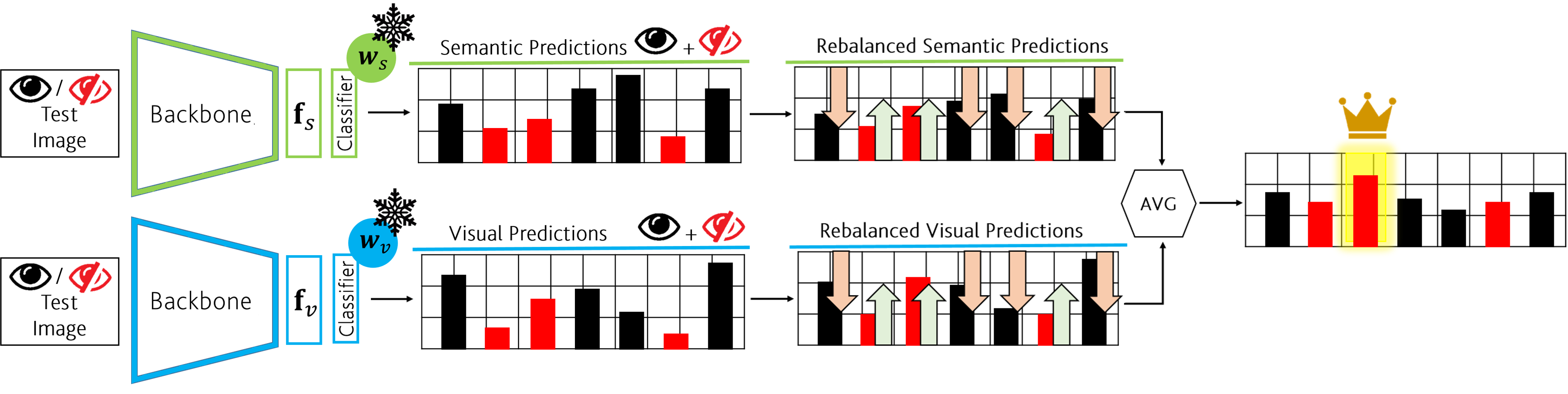}
    \caption{{\bf V\&S-CC-R: The Inference Stage.} With two, independent, forward passes, we feed any test image to a S-CC (green) and V-CC (blue) backbone, computing predictions. Predictions are smoothened unsing a temperature value $\tau$ (not shown here), rebalanced between seen (black) and unseen (red) using Eq. \ref{eq:modulated} and then combined with a plain averaging. Classes are still predicted using as single $\arg \max$ operation (selecting, in this case, the ``crowned'' class).}
    \label{fig:pipeline_inference}
\end{figure*}

In \cite{chao2016empirical}, this is loosely implemented by subtracting seen predictions scores by a fixed quantity $\gamma$ (calibrate stacking) while in \cite{atzmon2019COSMO} and adaptive prior is designed and rooted in a probabilistic framework of confidence smoothing (COSMO). Differently, in this paper we aim at pursuing a fully discriminative approach, and training a discriminator $D$ to predict whether a given test instance is seen or not. Thus, once we do so, we can 
%We empirically verify that, despite this actually happens, still the performance in standard ZSL is so solid that the relative state-of-the-art is improved by V\&S-CC (check Table \ref{tab:oracle}). This means that, despite our crafting is effectively transferrable onto unseen classes (otherwise, we won't be able to achieve SOTA performance in standard ZSL), the problem arises when comparing the predictor's scores on seen and unseen classes jointly. This is understandable given that we fine-tune the network only on seen classes only: overall, if an oracle selector is capable of telling us, for any test sample, whether it belongs to seen or unseen classes, we could totally circumvent this problem, we can largely outperform state-of-the-art, and even generative methods (see Table \ref{tab:soa}). Therefore, we propose to solve this issue by
re-modulate the estimated softmax probability scores $\mathbf{p}_j$ of our crafted network in the following manner
\begin{equation}\label{eq:modulated}
 \overline{\mathbf{p}}_j = \begin{cases} p_D \cdot \mathbf{p}_j & \mbox{if $j$ is a seen class,} \\
 (1-p_D) \cdot \mathbf{p}_j & \mbox{if $j$ is an unseen class.} \\
 \end{cases}
\end{equation}
In Eq. \eqref{eq:modulated}, $p_D$ is the probability estimated by $D$ for a test instance to belong to the seen pool of classes. That is, we seek to train $D$ to achieve the best approximation for an oracle, who would have given $p_D = 1$ for all seen test instances and $p_D = 0$ for all the unseen ones. 

%Differently, to \cite{chao2016empirical} we re-modulate probabilities, as opposed to adding/subtracting logits by a fixed offset hyper-parameter. Differently to the adaptive prior in \cite{atzmon2019COSMO}, we do not rely on a feature generator (as in \cite{atzmon2019COSMO}) to better pre-condition it, but we adopt a novel strategy based on dummy auxiliary data and mixup \cite{zhang2018mixup}: see the paragraph beneath.

\paragraph{Training the discriminator $D$.} We extract $p_D$ from a logistic regressor \cite{bishop2006pattern} trained with a binary cross-entropy loss to classify whether a given \textit{logits} is seen or not. The logits are obtained from our crafted convnet and we can exploit it to obtaining seen logits from seen training data. Since unseen classes are unavailable at training time, we need to \textit{synthesize unseen logits} and, to do so, we take advantage of the idea we visualize in Figure \ref{fig:rebalance}. 

To synthesize unseen logits, we consider task-irrelevant data from PASCAL VOC 2008 and we make sure to remove any overlapping class (\eg, removing animals when working on Animals with Attributes \cite{lampert2009learning}). We fed this task-irrelevant data into the network and extract logits from a trained model: \eg, we compute flowers-related logits for an image of a bag, see Fig. \ref{fig:rebalance}. Albeit surely not reliable in their predictions, we can still capitalize on those logits to increase the uncertainty of the model and simulate the out-of-distribution regime. But, since task-irrelevant data are surely bringing more uncertainty with respect to the one we would need for unseen data, we exploit mixup \cite{zhang2018mixup} (in the feature space) to mitigate this. We combine task-irrelevant logits with seen training one: that's how the computable surrogate logits are extracted and used as negative data to train $D$. %this will generate a surrogate for unseen logits which we cannot compute, so that we are still able to train $D$ in discriminating whether a given logit is seen or not, and implement the confindence rebalancing mechanism as in Eq. \eqref{eq:modulated}.

\paragraph{Observation.} Note that this rebalance mechanism is 1) only necessary for generalized ZSL and 2) can be combined with the semantic crafting S-CC, the visual crafting V-CC and the ensemble of the two (V\&S-CC) - see Fig. \ref{fig:pipeline_inference}.

\section{Experiments}\label{sez:exp}

\paragraph{Datasets and evaluation metrics} We run experiments on AWA \cite{AwithA}, CUB \cite{CUB}, SUN \cite{SUN}, FLO \cite{FLO} datasets using the proposed splits ``PS'' which are ImageNet-compatible \cite{xian2018zero}. Class embeddings are extracted using the TF-IDF \cite{Huang_2019_CVPR} for AWA, CUB, SUN and the LSTM-based approach of \cite{xian2019f-VAEGAN-D2} for FLO. Standard performance metrics are used \cite{xian2018zero}: in standard, inductive ZSL performance is evaluated as mean top-1 classification accuracy ($T1$) over unseen classes. In generalized, inductive ZSL, we report $H$, the harmonic mean between mean per-class accuracy scores $S$ and $U$, computed over seen or unseen classes independently \cite{xian2018zero}. %\textcolor{red}{I will also show results with human-annotated class embeddings for AWA2, CUB and SUN) and results on NAB/CUB in the setup of \cite{Huang_2019_CVPR}}.
%and NAB \cite{xian2018zero,Huang_2019_CVPR} using validating protocols that are (ImageNet compatible) and used from all prior works (``proposed splits'' \cite{xian2018zero} and SCS/SCE for NAB and CUB \cite{Huang_2019_CVPR}). See \textcolor{red}{additional details in the Supplementary Material} and check the beginning of Sec. \ref{sez:method} for the definition of the metrics $T1$ (ZSL) and $H$ and $\mathcal{A}$ (generalized ZSL) used here.

\begin{figure}[t!]
    \centering
    \vspace{12pt}
    \begin{overpic}[width=\columnwidth]{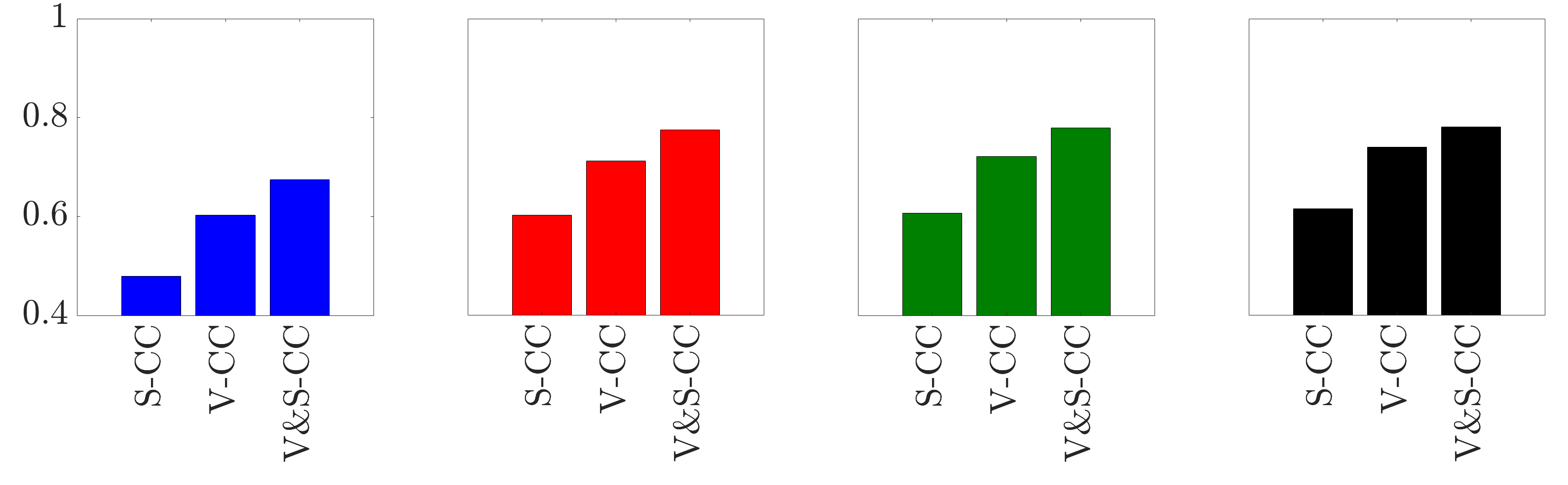}
    \put (6,31) {\rotatebox{0}{\bf \textcolor{blue}{AlexNet}}}
    \put(16,20) {\scalebox{0.6}{67.45\%}}
    \put (29.4,31) {\rotatebox{0}{\bf \textcolor{red}{ResNet-101}}}
    \put(41,23) {\scalebox{0.6}{77.51\%}}
    \put (53,31) {\rotatebox{0}{\bf \textcolor{green}{\scalebox{0.9}{DenseNet-201}}}}
    \put(65.8,23.4) {\scalebox{0.6}{77.92\%}}
    \put (80,31) {\rotatebox{0}{\bf \scalebox{0.9}{DarkNet-53}}}
    \put(90,23.4) {\scalebox{0.6}{78.10\%}}
    \end{overpic}
        \caption{{\em Different craftings for standard, inductive ZSL.} On FLO, using the $T1$ metric, we ablate the impact of different crafting mechanism, across different newtorks, once we craft them into a zero-shot learner}
    \label{fig:ablation_ZSL}
\end{figure}
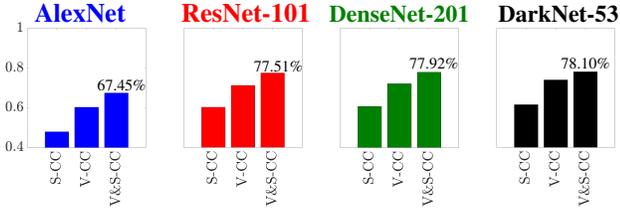

\paragraph{Turn your favorite convnet into a zero-shot learner!} Our idea of crafting the classifier's weights of a softmax operator to solve ZSL is broadly applicable to a number of deep convolutional networks, designed for image classification. In fact, in Fig. \ref{fig:ablation_ZSL}, we show how different networks -- in this case, AlexNet \cite{krizhevsky2012imagenet}, ResNet-101 \cite{he2016deep}, DenseNet-201 \cite{huang2017densely} and DarkNet-53 \cite{redomon2020darknet} (with 256 $\times$ 256 input layer) -- can be turned into a classifier that is able to discriminate classes that remain unseen at training time (on FLO using the V\&S-CC ensemble of visual and semantic crafting, we achieve $T1_{\rm AlexNet} = $~67.45\%, $T1_{\rm ResNet-101} = $~77.51\%, $T1_{\rm DenseNet-201} = $~77.92\% and $T1_{\rm DarkNet-53} = $~78.10\%).

\begin{figure}[t!]
    \centering
    \vspace{3pt}
    \textcolor{cc1}{\bf No rebalance}~~~~\textcolor{cc2}{\bf Cal. Stack}~\cite{chao2016empirical}~~~~\textcolor{cc3}{\bf COSMO }~\cite{atzmon2019COSMO}~~~~\textcolor{cc4}{\bf \textit{Ours}}
    \vspace{15pt}\\
    %\vspace{3pt}
    \begin{overpic}[width=\columnwidth]{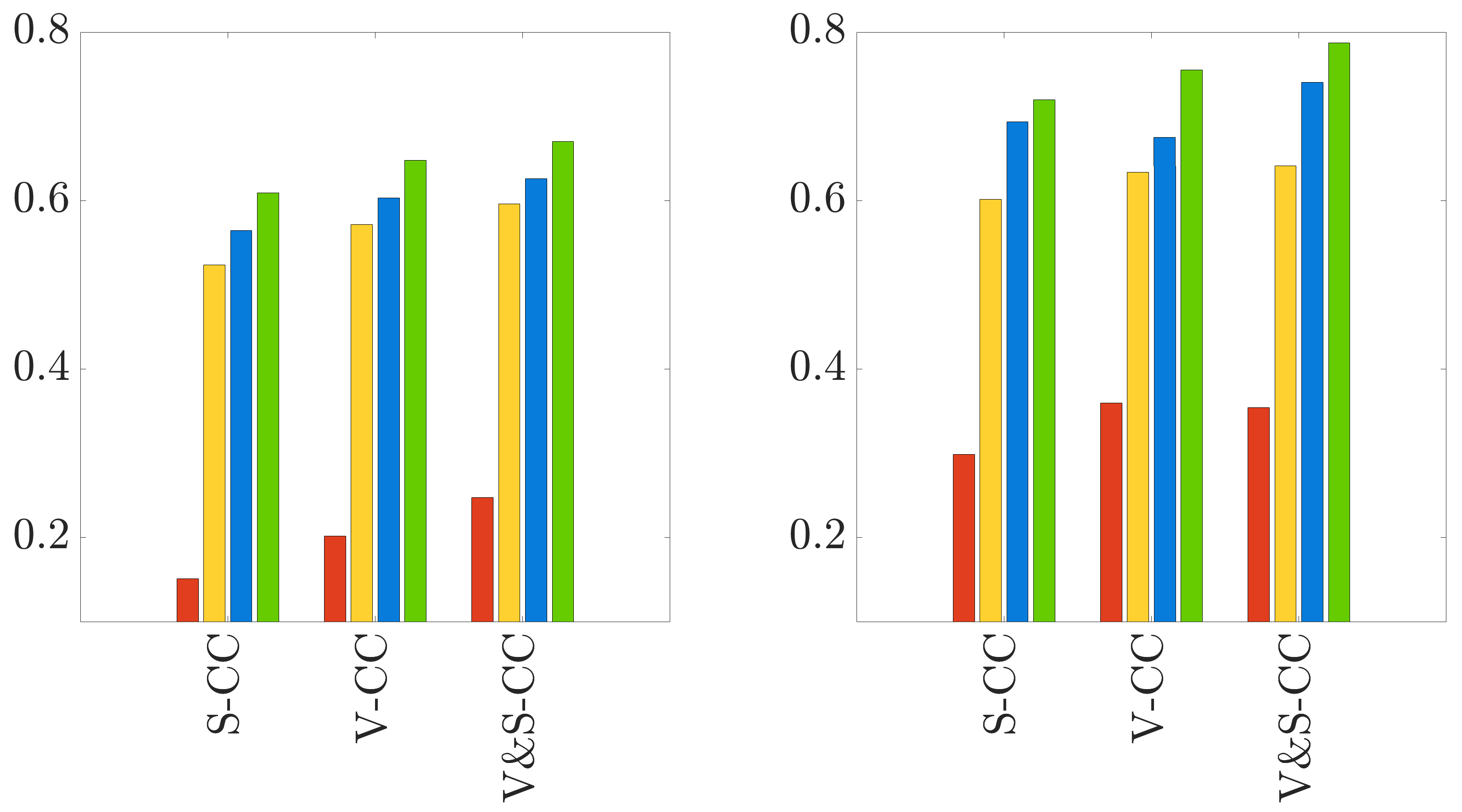}
    \put (6,54) {\bf AlexNet}
    \put (60,54) {\bf ResNet-101}
    \put (7,15) {\rotatebox{90}{GZSL, $H$ [\%]}}
    \put (60,15) {\rotatebox{90}{GZSL, $H$ [\%]}}
    \end{overpic}
    \includegraphics[width=\columnwidth]{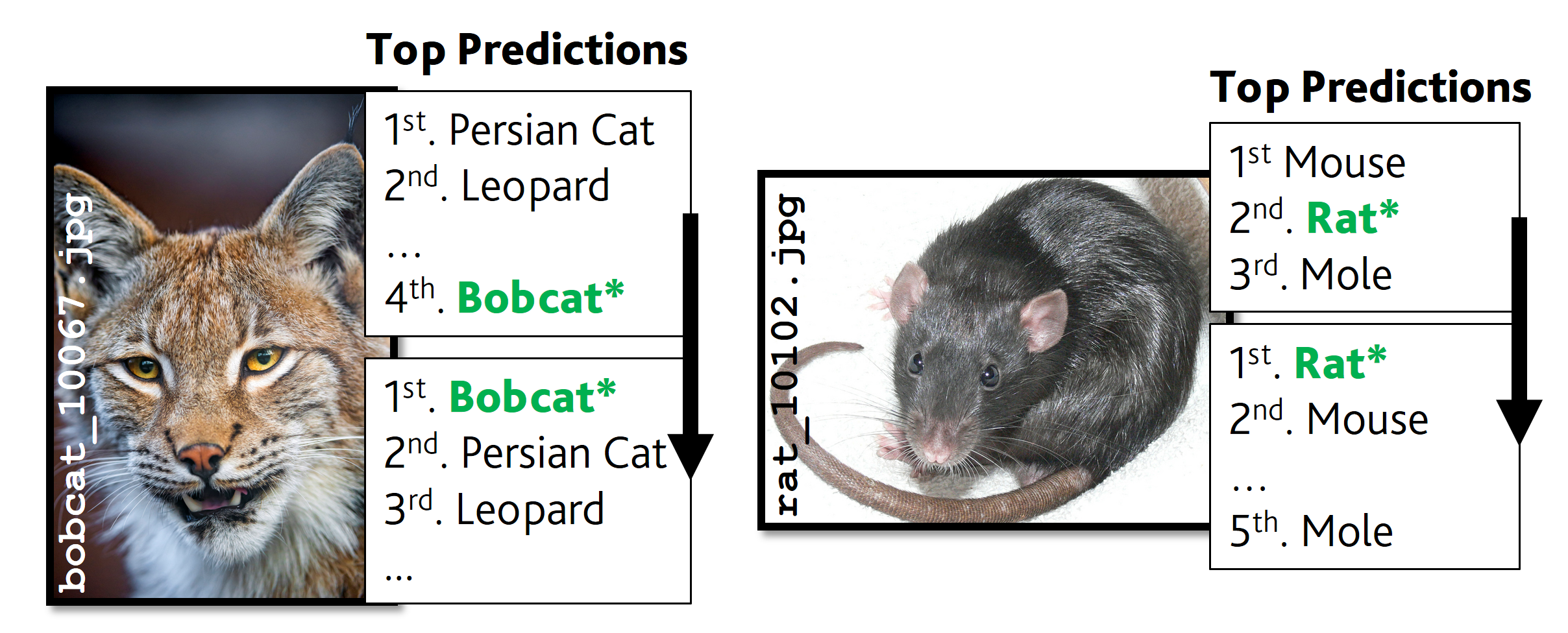}
    \caption{{\em Rebalancing in generalized, inductive ZSL.} \textit{Top}: On FLO, we ablate on different rebalancing methods applied to the AlexNet and ResNet-101 \textit{crafted} models. \textit{Bottom}: qualitative results before and after \textit{our} rebalancing (ground truth class in green, unseen classes are starred) on AWA. % See the Supplementary Material for additional results.
    }
    \label{fig:ablation_H}
\end{figure}

\paragraph{Ablation on predictions' rebalancing.} We compare our proposed confidence rebalancing mechanism (Section \ref{ssez:rebal}) with two standard, state-of-the-art alternatives: calibrate stacking \cite{chao2016empirical} and COSMO \cite{atzmon2019COSMO}. As the reader can see in Fig. \ref{fig:ablation_H} for FLO, our method is more effective than prior solutions  \cite{chao2016empirical,atzmon2019COSMO} in resolving the bias towards seen classes that are used to train S-CC, V-CC and V\&S-CC models, making them effective to generalize onto unseen classes also. %The trend is confirmed also on other datasets, as the reader can see in the Supplementary Material. 

% \begin{table}[h!]
% \centering
% \ra{1.3}
% \resizebox{\columnwidth}{!}{
% \begin{tabular}{@{}llllrlll@{}}\toprule
% & \multicolumn{3}{c}{ZSL: $T1$ [\%]} & ~ & \multicolumn{3}{c}{GZSL: $H$ [\%]} \\
% \cmidrule{2-4} \cmidrule{6-8}
% & S-CC & V-CC & V\&S-CC & ~ & S-CC & V-CC & V\&S-CC\\
% \cmidrule{2-4} \cmidrule{6-8}
% -- & 60.3 & 71.3 & 77.5 & ~ & 29.9 & 36.0 & 35.4 \\
% $R$ & $\rotatebox{45}{$=$} & $\rotatebox{45}{$=$} & $\rotatebox{45}{$=$} & ~ & 72.0 & 75.5 & 78.7\\
% $\boldsymbol{\mathcal{O}}$ & $\rotatebox{45}{$=$} & $\rotatebox{45}{$=$} & $\rotatebox{45}{$=$} & ~ & 73.8 & 82.1 & 86.9 \\
% \bottomrule
% \end{tabular}}
% \caption{Numerical comparison (on FLO, crafting a ResNet-101 model) on the effect of rebalancing ``R'' if compared to an orcale seen/unseen instance selector ``\mathcal{O}''.}
% \label{tab:oracle}
% \end{table}

On the same dataset, we provide a numerical ablation study to quantify the impact of performance of our rebalancing mechanism ``\textit{R}'' (across all different crafting strategies) while also comparing with an ``oracle'' discriminator $\mathcal{O}$ which is always able to spot whether a test instance is seen or not: as shown in Tab. \ref{tab:oracle}, the gap in between ``$R$'' and ``$\mathcal{O}$'' is only 1.8\% for S-CC, increases to 6.6\% for V-CC (but also the performance increases) and stabilizes to 8.1\%. Albeit improvable, this gap certifies that the discriminator is effective enough in handling unseen classes and make predictions more balanced: after adding ``$R$'', the performance in GZSL, measured through $H$, is always (more than) doubled. %See the Supplementary Material for a more extended table of results.

\begin{SCtable}[50]
%\begin{table}
\centering
\ra{1.3}
\resizebox{0.5\columnwidth}{!}{
\begin{tabular}{lccc}\toprule
& \multicolumn{3}{c}{GZSL: $H$ [\%]} \\
\cmidrule{2-4}
& S-CC & V-CC & V\&S-CC\\
\cmidrule{2-4}
& 29.9 & 36.0 & 35.4 \\
$R$ & 72.0 & 75.5 & 78.8\\
$\boldsymbol{\mathcal{O}}$ & 73.8 & 82.1 & 86.9 \\
\bottomrule
\end{tabular}}
\caption{Numerical {\bf ablation} (on FLO, crafting a ResNet-101 model) on the effect of rebalancing ``\textit{R}'' versus an oracle seen/unseen instance selector ``$\mathcal{O}$''.}
\label{tab:oracle}
\end{SCtable}

\paragraph{Learning a ZSL-tailored feature representation.} For V-CC, the crafting rules $r_i,\dots,\widetilde{r}_k,\dots$ are obtained by averaging seen features extracted from a pre-trained convnet (and learning the prototypes for the unseen). Hence, we can still perform ZSL inference by simply crafting the classifier, while skipping any fine-tuning on the feature representation $\mathbf{f}$. By doing so, we can really evaluate the benefit of tailoring visual descriptors for ZSL: it turns out that this is indeed advantageous, since the \textit{fine-tuning free variant of V-CC} is gapped in performance with respect to the full (\ie, fine-tuned) version (\eg, about -20\% on CUB). %A complete analysis is provided in the Supplementary Material.

%\sidecaptionvpos{table*}{c}
\begin{table*}%[][h!]
\centering
\ra{1.3}
\resizebox{0.85\textwidth}{!}{
\begin{tabular}{@{}rlrccccrcccc@{}}\toprule
& & & \multicolumn{4}{c}{Zero-Shot Learning $\colon T1$ [\%]} & \phantom{abc} & \multicolumn{4}{c}{Generalized ZSL $\colon H$ [\%]} \\
\cmidrule{4-7} \cmidrule{9-12}
& & \phantom{abc} & AWA & CUB & SUN & FLO & \phantom{abc} & AWA & CUB & SUN & FLO \\
\midrule
CondZSL & ${}^{\rm 2019}_{}~$\cite{li2019rethinking} && 71.1 & 54.4 & 62.6 & -- & \phantom{abc} & 66.7 & 47.5 & 39.3 & -- \\
LisGAN & ${}^{\rm 2019}_{}~$\cite{li2019leveraging} && 70.6	& 58.8 &	61.7 & 69.6 & \phantom{abc} & 62.3	& 51.6 &	40.2 & 68.3 \\
$f$-VAEGAN-$D2$ & ${}^{\rm 2019}_{}~$\cite{xian2019f-VAEGAN-D2} && 70.3 & 72.9 & 65.6 & 70.4  & \phantom{abc} & 65.2 & 68.9 & 43.1 & 75.1 \\
EXEM & ${}^{\rm 2020}_{}~$\cite{changpinyo2020classifier} && 68.1 & 58.6 & 62.9 & 69.4 & \phantom{abc} & 46.5 & 50.1 & 39.1 & 48.2 \\
ZSML &  ${}^{\rm 2020}_{}~$\cite{verma2020meta} && 76.1 & 69.6 & 60.2 & -- & \phantom{abc} & 65.8 & 55.7 & -- & -- \\
APN & ${}^{\rm 2020}_{}~$\cite{xu2020attribute} && 71.7 & 73.8 & 65.7 & -- & \phantom{abc} & 
65.6 & 70.0 & 43.7
& -- \\
ZS-OCD & ${}^{\rm 2020}_{}~$\cite{keshari2020generalized} && 71.3 & 	60.3 & 63.5	& -- & \phantom{abc} & 65.7	& 51.3 & 43.8 & -- \\
IZF & ${}^{\rm 2020}_{}~$\cite{shen2020invertible} && 74.5 & 67.1 & 68.4 & -- & \phantom{abc} & 68.0 & 59.4 & 54.8 & -- \\
$tf$-VAEGAN & ${}^{\rm 2020}_{}~$\cite{narayan2020latent} && 73.4 & 74.3 & 66.7 & 74.7 && 66.7 & 70.7 & 46.3 & {\bf 79.4} \\
AFRNet & ${}^{\rm 2020}_{}~$\cite{liu2020zero} && 75.1 & -- & 64.0 & -- & \phantom{abc} & 70.1 & -- & 41.5 & -- \\
CN & ${}^{\rm 2021}_{}~$\cite{sko2021norm} &&  -- & -- & -- & -- && 67.6 & 50.3 & 43.1 & -- \\
A\&G-ZSL & ${}^{\rm 2021}_{}~$\cite{chou2021adaptive} && 76.4 & 77.2	& 66.2 & -- & \phantom{abc} & 71.3 & 72.6 & 46.5 & -- \\
\midrule
\textbf{S-CC-(R)} & ${}^{\rm 2021}_{}~$\textbf{ours} && 71.7 & 79.7 & 63.5 & 60.3 && 70.0 & 70.5 & 55.0 & 72.0 \\
\textbf{V-CC-(R)} & ${}^{\rm 2021}_{}~$\textbf{ours} && 76.8 & 80.9 & 67.0 & 71.3 && 73.3 & 72.1 & 56.9 & 75.5 \\
\textbf{V\&S-CC-(R)} & ${}^{\rm 2021}_{}~$\textbf{ours} && \textbf{78.5} & \textbf{81.2} & \textbf{69.3} & \textbf{77.5} && \textbf{75.5} & \textbf{73.1} & \textbf{58.0} & 78.8 \\
%GZSL, AWA2 = u=67.0,s=86.5, H=75.5
%GZSL, FLO = u=0.7437    s=0.8380    H=0.7881
\bottomrule
\end{tabular}}
\caption{Comparison against recent state-of-the-art methods for standard and generalize, inductive ZSL. For fair comparisons, \textit{a ResNet-101 feature encoder is adopted since used in prior art}. While reporting either Top-1 classification accuracy $T1$ (ZSL) or the harmonic mean $H$ between unseen/seen mean per class accuracy (GZSL), the best performance is bolded. S-CC, V-CC and V\&S-CC are endowed of the rebalancing module ``R'' only for GZSL (in standard ZSL is simply unnecessary).} % Extended tables of results are available in the Supplementary Material.}
\label{tab:soa}
\end{table*}

\subsection{Comparison with SOTA in inductive (G)ZSL}
We compare our proposed visually and semantically crafting V\&S-CC against the state-of-the-art in inductive ZSL (standard and generalized protocols) by selecting a ResNet-101 backbone, since this common shared encoder is adopted from all prior art \cite{li2019rethinking,li2019leveraging,changpinyo2020classifier,verma2020meta,xu2020attribute,keshari2020generalized,shen2020invertible,narayan2020latent,liu2020zero,sko2021norm,chou2021adaptive}. We can therefore ensure a fair comparison - but note that we could have achieved slightly better results if using more recent architectures such as DarkNet-53 (see Fig. \ref{fig:ablation_ZSL}).

In details we compare with the conditional/probabilistic approaches of \cite{li2019rethinking,keshari2020generalized}, the syntesis of both classifier and exemples \cite{changpinyo2020classifier}, and a number of feature generating approaches using either GANs \cite{li2019leveraging,liu2020zero,chou2021adaptive}, VAE-GANs \cite{xian2019f-VAEGAN-D2,narayan2020latent} and GANs plus meta-learning \cite{verma2020meta}. We also compare against very recent works, relying on either invertible networks \cite{shen2020invertible} or feature scaling \cite{sko2021norm}. 

As show in Table \ref{tab:soa}, our method improved in performance ($T1$) previous methods in standard, inductive ZSL on AWA (\textit{+2.1}\% over \cite{chou2021adaptive}), CUB (\textit{+4.0}\% over \cite{chou2021adaptive}), SUN (\textit{+0.9}\% over \cite{shen2020invertible}) and FLO (\textit{+2.8}\% over \cite{narayan2020latent}). We also improve prior art in generalized, inductive ZSL on AWA (\textit{+4.2}\% over \cite{chou2021adaptive}) CUB (\textit{+0.5}\% over \cite{chou2021adaptive}) and SUN (\textit{+3.2}\% over \cite{shen2020invertible}). On FLO, we improve all prior methods except to tf-VAEGAN \cite{narayan2020latent}, only. We pay a gap of \textit{-0.6}\%: we deem the reason of that to be in the limited size of the dataset. In fact, tf-VAEGAN seems to resolve this problem by generating much more synthesized features if compared to the seen examples available (1200 generated descriptors per unseen class, while each seen class is represented only by one or two hundreds of examples).

%\textcolor{red}{Table \ref{tab:soa} and \ref{tab:oracle}. I need to discuss results \dots\texttt{WIP}.}

\begin{figure}[t!]
    \centering
    \begin{overpic}[width=\columnwidth]{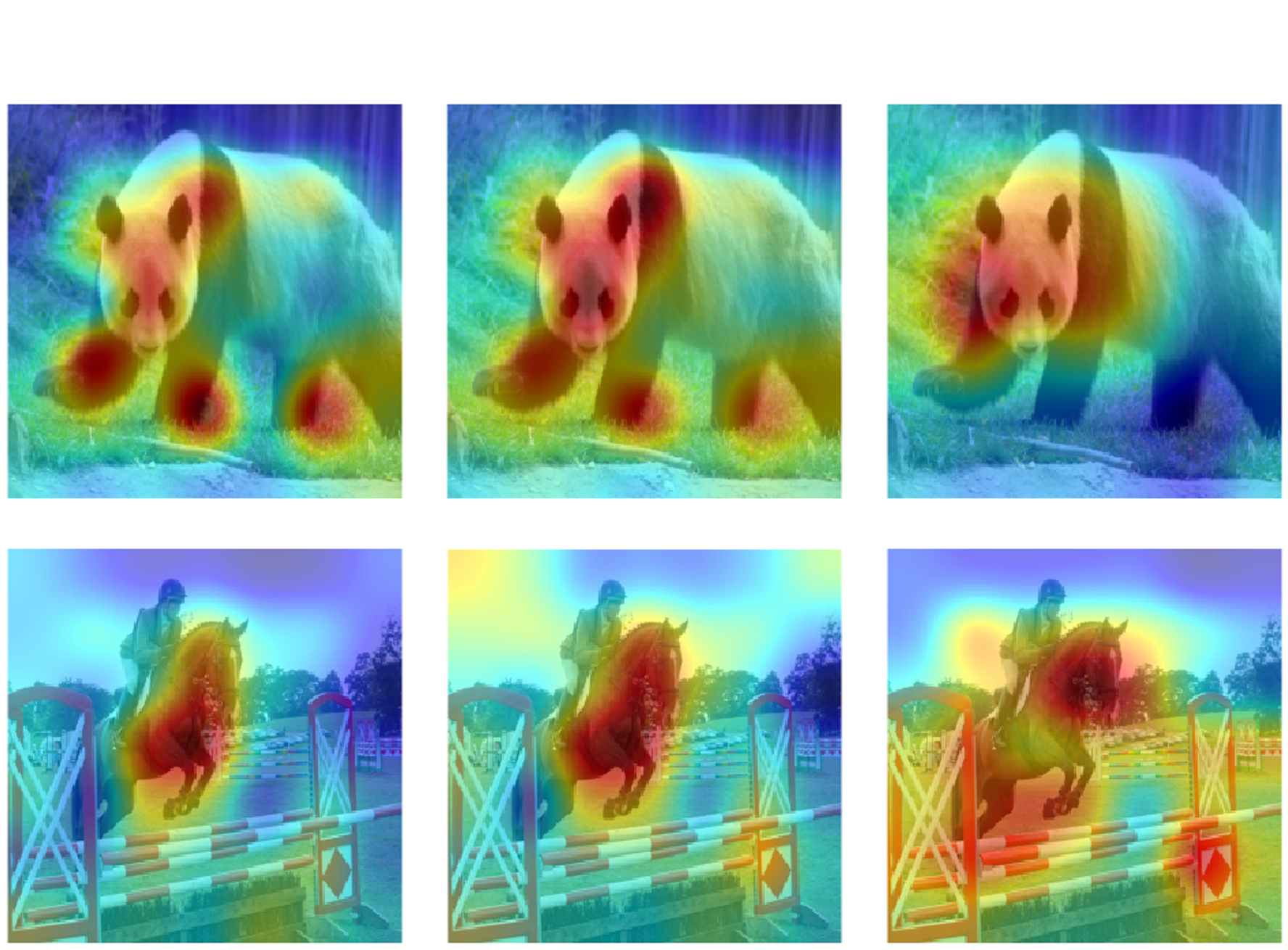}
    \put(1,67) {All attributes}
    \put(36,67) {\st{Is Timid}}
    \put(69,67) {\st{Is Black}}
    \end{overpic}
    \caption{Effect of removing non-visually grounded (\eg, is timid) and visually grounded (\eg, is black) attributes from class embeddings on Animals with Attributes \cite{AwithA}.}
    \label{fig:visualizations2}
\end{figure}

\paragraph{Interpretations with Saliency Maps.} To better ground why our recorder performance over prior art is superior, we exploited grad-CAM \cite{selvaraju2017grad} to inspect what happens when using manually annotated attributes' list for crafting on a S-CC model. As shown in Fig. \ref{fig:visualizations2}, when we compare the learnt representation on all attributes with the one obtained once we removed ``is timid'', we see very little variations: effectively such an attribute is not visually grounded and this result is actually understandable. At the same time, interestingly, if we remove the attribute ``is black'', then we observe much more changes in the grad-CAM maps (right-most column of Fig. \ref{fig:visualizations2}) since this attribute is instead visually grounded \cite{bustreo2019enhancing} and therefore it is actually impacting the learnt representation. We claim that this dependency between neural attention and attributes, can better justify why our proposed crafting is superior to prior methods, mainly because we better intertwine semantic and visual cues, and this is arguably a desirable proxy for an effective generalization towards unseen classes. %Additional visualizations are available in the Supplementary Material.

\section{Discussion and relationship with prior art}\label{sez:relwork}

\paragraph{(Generalized) Zero-Shot Learning.} In the seminal work \cite{lampert2009learning}, ZSL is framed as the regression problem of predicting attributes from visual data. Differently, we exploit seen class embeddings and seen images to obtain better features with which we can do zero-shot inference on unseen classes, using a softmax classifier. Due to the absence of feature learning in \cite{lampert2009learning}, attributes' predictions is highly suboptimal to our crafting - \eg, -32\% on AWA and -30\% on SUN). In  CondZSL \cite{li2019rethinking} and EXEM \cite{changpinyo2020classifier}, classifiers explicitly depends upon class embeddings, using a conditional/causal perspective or exploiting data self-representation, respectively. However, neither \cite{li2019rethinking} nor \cite{changpinyo2020classifier} perform feature learning from raw image data (they exploit pre-computed ImageNet-pretrained features) and, again they are suboptimal to us (V\&S-CC always improve in performance both CondZSL nd EXEM, as one can check in Table \ref{tab:soa}).

\textit{End-to-end learnable} zero-shot learning methods are emerging as recent directions for both video-based \cite{brattoli2020rethinking} (not considered here) or image-based applications \cite{li2018deep}. However, in \cite{li2018deep}, the absence of a confidence rebalancing mechanism forbid the methods to be tested in GZSL - while in standard ZSL we are on par with \cite{li2018deep} on AWA (see \cite[Tab. 2]{li2018deep}), superior on SUN (+2.3\%) and sharply superior on CUB (+15\%). Note that we did not insert \cite{li2018deep} in Table \ref{tab:soa} since it did not provide results using ResNet-101 features, using GoogleNet features instead. Note that, in fact, we could have furthermore improved our performance in Table \ref{tab:soa} is using DarkNet-53 as opposed to ResNet-101 (\eg, about +3\% improvement for $H$ on FLO), but this would not have been fair with respect to prior art. \textit{Ensemble approaches} have been recently investigated in the (easiest) transductive ZSL setup - where unannotated visual data from unseen classes are accessible during training - \cite{felix2019multimodal,ye2019progressive}. To the best of our knowledge we are the first of proposing an ensemble in inductive (standard and generalized) ZSL. 

There are ZSL methods which, similarly to us, explicitly achieve \textit{interpretability} using class attention maps: \cite{Chen2020Ensemble,xu2020attribute}. Please, note that, in \cite{Chen2020Ensemble}, an explicit patch-based optimization is designed, while \cite{xu2020attribute} exploits body parts annotations (and not all ZSL datasets are provided of that and, usually this type of meta-data is not adopted). Our work is different since we achieve interpretability at not additional cost (no direct optimization for that, no extra annotation is required beyond class embeddings).

\paragraph{\bf Classifiers' Crafting.} We briefly discuss methods which combine a fixed classifier with a learnable feature representation. There is a recent sequence of works \cite{qi2018low,hoffer2018fix,pernici2019maximally,pernici2020class} applying this idea to diverse applications, namely, vanilla classification \cite{hoffer2018fix,pernici2019maximally}, incremental learning \cite{pernici2020class} or few-shot learning. To the best of our knowledge, our work is the first one to progress this sequence, applying this paradigm in inductive zero-shot learning. %Differently to \cite{hoffer2018fix,pernici2019maximally,pernici2020class,qi2018low}, we are not allowed to use even a single instance of some of the classes to recognize (the unseen), but we're still asked to generalize on them.

In \cite{hoffer2018fix,pernici2019maximally,pernici2020class}, the classifier's weights are fixed to a pre-defined geometrical configuration (for instance, mapping all classes to be recognized over the vertex of a simplex \cite{pernici2019maximally}). We cannot access all classes to be recognized at training time, but only the seen. Thus, we cannot adapt on unseen using newly incoming data from them (as in \cite{pernici2020class}) but we need to carry out predictions from zero shots. Hence, the classifier's weights we craft cannot be explicitly dependent upon unseen data - and this is different from the few-shot learning approach of \cite{qi2018low} in which data from all classes are used for this. We therefore opted for a solution in which we rely on the intrinsic seen-to-unseen transferrability of our fixed classification rules (used for crafting) so that the representation, that we learn by matching them, can be transferrable in turn.

\paragraph{\bf Recalibrated Predictions.} As shown in Fig. \ref{fig:ablation_H}. %and further expanded in the Supplementary Material, 
our proposed discriminator, predicting if a logit is seen or not while training on real seen logits and fake auxiliary unseen ones, seems valid in re-calibrating predictions in the generalized zero-shot learning regime (since enhancing the performance in GZSL quite sharply). Relatively to calibrate stacking \cite{chao2016empirical} or confidence smoothing \cite{atzmon2019COSMO}, it appears more effective in this respect. Beyond performance, we also technical differ from these works: our rebalancing is fully discriminative, while  \cite{chao2016empirical} is based on thresholding (by manually decreasing the predictor's confidence) and \cite{atzmon2019COSMO} seeks for an adaptive probabilistic prior.

%\paragraph{\bf Mixup.}
%Mixup is a popular data augmentation technique based on taking convex combinations of pairs of examples and their labels. Originally performed in the feature space \cite{zhang2018mixup}, it has been also demonstrated beneficial effects while applying to the input space directly, generating convex combinations of images for semi-supervised learning \cite{wang2019semi}. This simple technique has been shown to substantially improve both the robustness and the generalization of the trained model \cite{thulasidasan2019mixup,arazo2019unsupervised}.

\section{Conclusion}\label{sez:conclusions}

Our papers shows that we can turn a convnet into a zero-shot learner by crafting the weights of the softmax operator, using fixed semantic/visual classification rules. We show that this strategy, when combined with an ensemble, and furthermore boosted by a predictions' rebalancing, outperforms prior art in inductive zero-shot learning, on standard and generalized evaluation protocols, respectively.
Since using a convnet for ZSL inference, we can achieve an interpretable predictor, at no additional cost, through neural attention by using methods such as GRAD-CAM as they are.

% can combine them with a plain vanilla ensemble and outperform pior art on standard, inductive zero-shot learning on AWA, CUB, SUN and FLO datasets. While the predictions of our models are rebalanced (actually, re-modulated) using the predictions of a seen vs. unseen discriminator, we can outperform prior generalized zero-s

%\textcolor{red}{Limitations: two forward passes as opposed to one only to do inference\dots\texttt{WIP}}

%\section*{Acknowledgements} We are thankful to Valentina Sanguineti and, especially, Shah Nawaz for their proof-reading, to Davide Talon and Andrea Zunino for the useful discussion on neural saliency maps, and to Jacopo Tessadori a reproducibility test on our code.

{\small
\bibliographystyle{ieee_fullname}
\bibliography{egbib} }
\balance

\end{document}